%% file: newton-cg.tex
\documentclass[]{article}

\usepackage{amssymb,amsfonts}
\usepackage{graphicx}
\usepackage{textcomp}
\usepackage{xcolor}

\usepackage[linesnumbered,ruled,vlined,onelanguage]{algorithm2e}
\usepackage{amsmath}

\usepackage{url}
\usepackage{color}
\usepackage{verbatim}

\usepackage{float} 
\usepackage{multirow}

\usepackage{amssymb}
\usepackage{comment}
\input{newcmnds.tex}

\def\BibTeX{{\rm B\kern-.05em{\sc i\kern-.025em b}\kern-.08em
    T\kern-.1667em\lower.7ex\hbox{E}\kern-.125emX}}

\usepackage{hyperref}
\hypersetup{
	colorlinks   = true,
	linkcolor    = blue,
	citecolor    = green
}

\usepackage{physics}
\usepackage{subcaption}

\author{
Chih-Hao Fang
\thanks{Comp. Sci. Dept
Purdue Univ., W. Lafayette,
Indiana 47907, US
\tt fang150@purdue.edu}
\and
Sudhir B. Kylasa
\thanks{Elec. and Comp. Engg. Dept
Purdue Univ., W. Lafayette,
Indiana 47907, US
\tt skylasa@purdue.edu}
\and
Fred Roosta
\thanks{School of Mathematics and
 Physics, Univ. of Queensland 
St Lucia, QLD 4072, Australia
\tt fred.roosta@uq.edu.au}
\and
Michael W. Mahoney
\thanks{ICSI and Department of Statistics 
Univ. of California at Berkeley 
Berkeley, CA 94720, US
\tt mmahoney@stat.berkeley.edu}
\and
Ananth Grama
\thanks{Comp. Sci. Dept
Purdue Univ., W. Lafayette,
Indiana 47907, US 
\tt ayg@cs.purdue.edu}
}

\begin{document}

\title{Newton-ADMM: A Distributed GPU-Accelerated Optimizer for Multiclass Classification Problems}

\maketitle

\input{abstract.tex}
\input{intro.tex}
\input{related.tex}
\input{model.tex}

\input{expt.tex}

\input{conclusions.tex}

\(\)


\input{supplementary}
\end{document}

%% file: newcmnds.tex
\usepackage{enumitem}
\setlist[enumerate,1]{leftmargin=*,wide=0em, noitemsep,nolistsep, label = {\bfseries \arabic*.}}
\setlist[itemize,1]{leftmargin=*,wide=0em, noitemsep,nolistsep}

\newcommand {\zz}  { {\bf z} }
\newcommand {\bgg}  { {\bf g} }

\newcommand {\xx}  { {\bf x} }
\renewcommand {\aa}  { {\bf a} }

\newcommand {\yy}  { {\bf y} }

\newcommand {\rr}  { {\bf r} }

\newcommand {\HH}  { {\bf H} }

\DeclareMathOperator*{\argmin}{arg\,min}

\newcommand {\pp}  { {\bf p} }

\newcommand {\vv}  { {\bf v} }

\newcommand {\dd}  { {\bf d} }
\newcommand {\ee}  { {\bf e} }

\renewcommand{\vec}[1]{\ensuremath{\mathbf{#1}}}

\newcommand{\A}{\vec{A}}
\renewcommand{\H}{\vec{H}}

\newcommand{\V}{\vec{V}}
\newcommand{\W}{\vec{W}}

\newcommand{\defeq}{\mathrel{\mathop:}=}

\renewcommand{\Pr}{\hbox{\bf{Pr}}}

\definecolor{forestgreen}{rgb}{0.13, 0.55, 0.13}

\newcounter{comment}

\newenvironment{remark}[1][Remark]{\begin{trivlist}
\item[\hskip \labelsep {\bfseries #1}]}{\end{trivlist}}

\newcommand*\lin[1]{\langle #1\rangle}

%% file: abstract.tex
\begin{abstract}
 First-order optimization methods, such as stochastic gradient descent (SGD) and its variants, are widely used in machine learning applications due to their simplicity and low per-iteration costs. However, they often require larger numbers of iterations, with associated communication costs in distributed environments. In contrast, Newton-type methods, while having higher per-iteration costs, typically require a significantly smaller number of iterations, which directly translates to reduced communication costs.

In this paper, we present a novel distributed optimizer for classification problems, which integrates a GPU-accelerated Newton-type solver with the global consensus formulation of Alternating Direction of Method Multipliers (ADMM). By leveraging the communication efficiency of ADMM, GPU-accelerated inexact-Newton solver, and an effective spectral penalty parameter selection strategy, we show that our proposed method (i) yields better generalization performance on several classification problems; (ii) significantly outperforms state-of-the-art methods in distributed time to solution; and (iii) offers better scaling on large distributed platforms.

\end{abstract}

%% file: intro.tex
\section{Introduction}
Estimating parameters of a model from a given dataset is a critical component of a wide variety of machine learning (ML) applications. 
The parameter estimation problem often translates to one of finding a minima of a suitably formulated objective function. The key challenges in modern ``big-data'' problems relate to very large numbers of model parameters (which translates to high dimensional optimization problems), large training sets, and learning models with low generalization errors. Recognizing the importance of the problem, a significant amount of research effort has been invested into addressing these challenges.

The most commonly used optimization technique in machine learning is gradient descent and its stochastic variant, stochastic gradient descent (SGD). 
Gradient descent algorithms, which solely rely on gradient information, are often referred to as first-order methods. Recent results \cite{kylasa2018gpu,berahas2017investigation,roosta2019sub} have shown that the use of curvature information in the form of Hessian, or approximations thereof, can lead to significant improvements in terms of performance as manifest in their convergence rate, time, and quality of solutions.

A key challenge in optimization for machine learning problems is the large, often, distributed nature of the training dataset. It may be infeasible to collect the entire training set at a single node and process it serially because of resource constraints (the training set may be too large for a single node, or that the associated data transfer overhead may be large), privacy  (data may be constrained to specific locations), or the need for reducing optimization time. In each of these cases, there is a need for optimization methods that are suitably adapted to parallel and distributed computing environments.

Distributed optimization solvers adopt one of two strategies -- (i) executing each operation in conventional solvers (e.g., SGD or (quasi) Newton) in a distributed environment \cite{chen2016revisiting,goyal2017accurate,jin2016scale,dean2012large,wang2017giant,crane2019dingo,daneshmand2016dynaNewton,reddi2016aide,zhang2015disco}; or (ii) executing an ensemble of local optimization procedures that operate on their own data, with a coordinating procedure that harmonizes the models over iterations \cite{xu2017adaptiveb,xu2017adaptivea}.  The trade-offs between these two methods are relatively well understood in the context of existing solvers -- namely that the communication overhead of methods in the first class is higher, whereas, the convergence rate of the second class of methods is compromised. For this reason, methods in the first class are generally preferred in tightly coupled data-center type environments, whereas methods in the latter class are preferred for wide area deployments.

Alternating Direction Method of Multipliers (ADMM), is a well known method in 
distributed optimization for solving consensus problems \cite{boyd2011distributed}. 
To achieve superior convergence and efficient solution of the corresponding sub-problems, the choices of penalty parameters associated with global consensus and inner (node-local) sub-problem solver are critical. In particular, the quality of inner sub-problem solutions dictates the accuracy of the descent direction computed by ADMM. In this paper, we present a novel solver that uses the Spectral Penalty Selection (SPS) method in ADMM \cite{xu2017adaptiveb} for setting the penalty parameters and a variant of Newton's method as sub-problem solver. 
Our choices are motivated by the observation that first-order solvers are known to suffer from slow convergence rates, and are notoriously sensitive to problem ill-conditioning and the choice of hyper-parameters. In contrast, Newton-type methods are less sensitive to such adversarial effects. However, this feature comes with increased per-iteration computation cost. In our solution, we leverage lower iteration counts to minimize communication cost and efficient GPU implementations to address increased computational cost.


\paragraph{\textbf{Contributions:}} 
Our contributions in this paper can be summarized as follows:
\textit{
\begin{itemize}
\item We propose a novel distributed,
GPU-accelerated Newton-type method based on an ADMM framework that has low communication overhead, good per-iteration compute characteristics through effective use of GPU resources, superior convergence properties, and minimal resource overhead. 
\item Using a range of real-world datasets (both sparse and dense), we demonstrate that our proposed method yields significantly better results compared to a variety of state-of-the-art distributed optimization methods. 
\item Our pyTorch implementation is publicly available\footnote{https://github.com/fang150/Newton\_ADMM}. Our solver can be readily used for practical applications by data scientists and can be easily integrated with other well-known tools like Tensoflow.
\end{itemize}}

%% file: related.tex
\section{Related Research}
\label{sec:related-work}
First-order methods \cite{beck2017first,bubeck2015convex} -- gradient descent and its variants are commonly used in ML applications. This is mainly because these methods are simple to implement and have low per-iteration costs. However,  it is known that these methods often take a large number of iterations to achieve reasonable generalization, primarily due to their sensitivity to problem ill-conditioning. 
Second-order methods make use of curvature information, in the form the Hessian matrix, and as a result are more robust to problem ill-conditioning \cite{roosta2019sub}, and hyper-parameter tuning \cite{berahas2017investigation,xuNonconvexEmpirical2017}. However, they can have higher memory and computation requirements due to the application of the Hessian matrix. In this context, quasi-Newton methods \cite{nocedal2006numerical} can be used to approximate the Hessian by using the history of gradients. 
However, a history of gradients must be stored in order to approximate the Hessian matrix, and extra computation cost is incurred to satisfy the strong Wolfe condition. In addition, these methods are observed to be unstable when used on mini-batches \cite{kylasa2018gpu}. 

Several distributed solvers have been developed recently \cite{chen2016revisiting,goyal2017accurate,jin2016scale,dean2012large,wang2017giant,crane2019dingo,daneshmand2016dynaNewton,reddi2016aide,zhang2015disco}. Among these, \cite{chen2016revisiting,goyal2017accurate,jin2016scale,dean2012large} are classified as first-order methods. Although they incur low computational overhead, they have higher communication costs due to a large number of messages exchanged per mini-batch and high total iteration counts. Second-order variants \cite{wang2017giant,crane2019dingo,daneshmand2016dynaNewton,reddi2016aide,zhang2015disco} are designed to improve convergence rate, as well as to reduce  communication costs. DANE \cite{daneshmand2016dynaNewton}, and the accelerated scheme AIDE \cite{reddi2016aide} use SVRG \cite{johnson2013accelerating} as the subproblem solver to approximate the Newton direction. These methods are often sensitive to the fine-tuning of SVRG. DiSCO \cite{zhang2015disco} uses distributed preconditioned conjugate gradient (PCG) to approximate the Newton direction. The number of communications across nodes per PCG call is proportional to the number of PCG iterations. In contrast to DiSCO, GIANT \cite{wang2017giant} executes CG at each node and approximates the Newton direction by averaging the solution from each CG call. Empirical results have shown that GIANT outperforms DANE, AIDE, and DiSCO. The solver of Dunner et al. \cite{dunner2018distributed} is shown to outperform GIANT, however, it is constrained to sparse datasets. A recently developed variant, DINGO \cite{crane2019dingo}, can be applied to a class of non-convex functions, namely invex \cite{craven1981invex}, which includes convexity as a special sub-class. However, in the absence of invexity, the method can converge to undesirable stationary points. 

A popular choice in distributed settings is ADMM \cite{boyd2011distributed}, which combines dual ascent method and the method of multipliers. ADMM only requires one round of communication per iteration. However, ADMM's performance is greatly affected by the selection of the penalty parameter \cite{xu2017adaptiveb,xu2017adaptivea} as well as the choice of local subproblem solvers. 

%% file: model.tex
\section{Problem Formulation and Algorithm Details}
\label{sec:theory}

In this section, we describe the optimization problem formulation, and present our proposed Newton-ADMM optimizer. 

\paragraph{\textbf{Notation}}
We use bold lowercase letters to denote vectors, e.g., $\vv$, and bold upper case letters to denote matrices, e.g., 
$ \V $. $\nabla f(\xx)$ and $\nabla^{2} f(\xx)$ represent the gradient and the Hessian of function $f$ at $\xx$,
respectively. The superscript, e.g., $\xx^{(k)}$, denotes iteration count, and the subscript, e.g., $ \xx_{i} $, 
denotes the \textit{local}-value of the vector $ \xx $ at the $ i^{th} $ compute node in a distributed setting. 
$\mathcal{D}$ denotes the input dataset, and its cardinality is denoted by $|\mathcal{D}|$. 
Function $ F_i(\xx) $ represents the objective
function, $ F(\xx) $, evaluated at point $ \xx $ using $ i^{th}-$ observation. Function $ F_{\mathcal{D}}(\xx) $
represents the objective function evaluated on the entire dataset $\mathcal{D}$. 




\subsection{Problem Formulation} \label{sec:multi_class}

Consider a finite sum optimization problem of the form: 
\begin{align}
\label{eq:obj}
\min_{\xx \in \mathbb{R}^d} F(\xx) \triangleq \sum_{i=1}^n f_i(\xx) + g(\xx),
\end{align}
where each $f_{i}(\xx)$ is a smooth convex function and $ g(\xx) $ is a (strongly) convex and smooth regularizer. In ML applications, $f_{i}(\xx)$ can be viewed as loss (or misfit) corresponding to the $i^{th}$ observation (or measurement)\cite{tibshirani1996regression, friedman2001elements,bottou2016optimization, sra2012optimization}. In our study, we choose multi-class classification using soft-max and cross-entropy loss function, as an important instance of finite sum minimization problem. Consider a $p$
dimensional feature vector $\aa$, with corresponding labels $b$, drawn from $C$ classes. In such a classifier, the probability that $\aa$ belongs to a
class $c \in \{1,2,\ldots,C\}$ is given by:
\begin{align*}
\Pr\left(b = c \mid \aa,\xx_{1},\ldots, \xx_{C}\right) = \frac{e^{\lin{\aa,\xx_{c}}}}{\sum_{c' = 1}^{C} e^{\lin{\aa, \xx_{c'}}}},
\end{align*}
where $ \xx_{c} \in \mathbb{R}^{p}$ is the weight vector corresponding to class $ c $.
Recall that there are only $ C-1 $ degrees of freedom, since probabilities must
sum to one. Consequently, for training data
$\{\aa_{i},b_{i}\}_{i=1}^{n} \subset \mathbb{R}^{p} \times \{1,2,\ldots,C\}$,
the cross-entropy loss function for $ \xx = [\xx_{1}; \xx_{2}; \ldots; \xx_{C-1}] \in \mathbb{R}^{(C-1)p} $
can be written as:
\begin{align}
\label{eq:softmax_log_likelihood}
&F(\xx) \triangleq F(\xx_{1},\xx_{2},\ldots,\xx_{C-1}) \nonumber \\
= &\sum_{i=1}^{n} \left(\log \left(1+\sum_{c' = 1}^{C-1} e^{\lin{\aa_{i}, \xx_{c'}}}\right) - \sum_{c = 1}^{C-1}\mathbf{1}(b_{i} = c) \lin{\aa_{i},\xx_{c}}\right).
\end{align} 
Note that  $ d = (C-1) p $.  After the training phase, a new data instance $\aa$ is classified as: 
\begin{align*}
b = \arg \max &\left\{\left\{\frac{e^{\lin{\aa,\xx_{c}}}}{\sum_{c' = 1}^{C-1} e^{\lin{\aa, \xx_{c'}}}}\right\}_{c=1}^{C-1}, 1- \frac{e^{\lin{\aa,\xx_{1}}}}{\sum_{c' = 1}^{C} e^{\lin{\aa, \xx_{c'}}}}  \right\}.
\end{align*}

\subsection{Numerical Stability}
\label{sec:numerical_stability}

To avoid over-flow in the evaluation of exponential functions 
in~\eqref{eq:softmax_log_likelihood}, we use the ``Log-Sum-Exp''
trick~\cite{murphy2012machine}. Specifically, for each data point $\aa_{i}$, we first find
the maximum value among $\lin{\aa_{i},\xx_{c}}, \; c=1,\ldots, C-1$. Define:
\begin{align}
\label{eq:max_x}
M(\aa) = \max \Big\{ 0, \lin{\aa,\xx_{1}}, \lin{\aa,\xx_{2}}, \ldots, \lin{\aa,\xx_{C-1}} \Big\}, 
\end{align}
and 
\begin{align}
\label{eq:log_sum_trick}
\alpha(\aa) \defeq e^{-M(\aa)}+\sum_{c' = 1}^{C-1} e^{\lin{\aa, \xx_{c'}} - M(\aa)}.
\end{align}

Note that $M(\aa) \geq0, \alpha(\aa) \geq 1$. Now, we have:
\begin{align*}
1+\sum_{c' = 1}^{C-1} e^{\lin{\aa_{i}, \xx_{c'}}} 
& = e^{M(\aa_{i})} \alpha(\aa_{i}).
\end{align*}
For computing~\eqref{eq:softmax_log_likelihood}, we use:
\begin{align*}
& \log \left(1+\sum_{c' = 1}^{C-1} e^{\lin{\aa_{i}, \xx_{c'}}}\right) \\
&= M(\aa_{i}) + \log \left(e^{-M(\aa_{i})}+\sum_{c' = 1}^{C-1} e^{\lin{\aa_{i}, \xx_{c'}} - M(\aa_{i})}\right) \\
&= M(\aa_{i}) + \log \big( \alpha(\aa_{i}) \big).
\end{align*}
Note that in all these computations, we are guaranteed to have all the exponents appearing
in all the exponential functions to be negative, hence avoiding numerical over-flow.

\subsection{ADMM Framework}

Let $ \mathcal{N} $ denote the number of nodes (compute elements) in the distributed environment. 
Assume that the input dataset $ \mathcal{D} $ is split among the $ \mathcal{N} $ nodes as $ \mathcal{D} = \mathcal{D}_{1} \cup \mathcal{D}_{2} \ldots \cup \mathcal{D}_{\mathcal{N}} $. 
Using this notation, \eqref{eq:obj} can be written as:
\begin{align}
\label{eq:dist-obj}
\min & \sum_{i = 1}^{\mathcal{N}}  \sum_{j \in \mathcal{D}_i} f_j( \xx_i ) + g( \zz )   \\
& \textrm{s.t.} \quad  \xx_i - \zz = 0 , \quad  i = 1, \ldots, \mathcal{N}, \nonumber 
\end{align}
where $ \zz $ represents a global variable enforcing consensus among $\xx_{i}$'s at all the nodes. 
In other words, the constraint enforces a consensus among the nodes so that all the local variables, 
$ \xx_i $, agree with global variable $\zz$. The formulation \eqref{eq:dist-obj} is often referred to as a \textit{global consensus} problem.  ADMM is based on an augmented Lagrangian framework; it solves the global consensus problem by alternating iterations on primal/ dual variables. In doing so, it inherits the benefits of the decomposability of dual ascent and the superior convergence properties of the method
of multipliers. 

ADMM methods introduce a penalty parameter $\rho$, which is the weight on the measure of \textit{disagreement} between $ \xx_{i} $'s and 
global consensus variable, $ \zz $. The most common adaptive penalty parameter selection is 
Residual Balancing \cite{boyd2011distributed}, which tries to balance the 
dual norm and residual norm of ADMM. 
Recent empirical results using SPS \cite{xu2017adaptiveb}, which is based on the estimation of the local curvature of subproblem on each node, yields significant improvement in the efficiency of ADMM. Using the SPS strategy for penalty parameter selection, ADMM iterates can be written as follows: 
{\small
\begin{subequations}
\begin{align}
\xx_{i}^{k+1} &= \argmin_{\xx_i} f_i(\xx_i) + \frac{\rho_i^{k}}{2} || \zz^{k} - \xx_i + \frac{\yy_{i}^{k}}{\rho_i^{k}} ||_{2}^{2}, \label{eq:x-update} \\
\zz^{k+1} &= \argmin_{\zz} g(\zz) + \sum_{i=1}^{\mathcal{N}}  \frac{\rho_i^{k}}{2} || \zz -\xx_{i}^{k+1} +\frac{\yy_{i}^{k}}{\rho_i^{k}}  ||_{2}^{2}, \label{eq:z-update} \\
\yy_{i}^{k+1} &= \yy_{i}^{k} + \rho_i^{k} (\zz^{k+1}- \xx_i^{k+1} ). \label{eq:y-update}
\end{align}
\end{subequations}
}%
With $\ell_{2}-$regularization, i.e., $ g(\xx) = \lambda \|\xx\|^{2}/2 $, \eqref{eq:z-update} has a closed-form solution given by
\begin{align}
\zz^{k+1}(\lambda + \sum_{i=1}^{\mathcal{N}}\rho_i^{k}) = \sum_{i=1}^{\mathcal{N}} \bigl[\rho_i^{k} \xx_i^{k+1}- \yy_i^{k}  \bigr], 
\end{align}
where $\lambda$ is the regularization parameter.

Algorithm \ref{alg:dist-Newton} presents our proposed method incorporating the above formulation of ADMM.

\begin{algorithm} [!htbp]
	\caption{ADMM method (outer solver)}
	\label{alg:dist-Newton}
	\SetAlgoLined
	\SetKwInOut{Input}{Input}
	\SetKwInOut{Parameter}{Parameters}
	\Input{$\xx^{(0)}$ (initial iterate),  $\mathcal{N}$ (no. of nodes) }
	\Parameter{ $ \beta $, $ \lambda $ and $ \theta < 1$}
	Initialize $\zz^{0}$ to 0 \label{dn:line:1} \\
    Initialize $\yy_i^{0}$ to 0 on all nodes.\label{dn:line:2} \\
	\ForEach{$k = 0,1,2,\ldots$}{
	 Perform Algorithm \ref{alg:standalone-Newton} with, $\xx_i^{k}$, $\yy_i^{k}$, and $\zz^{k}$ on all nodes \\
	 Collect all local $\xx_i^{k+1} $ \\
		 Evaluate $\zz^{k+1}$ and $\yy_i^{k+1}$ using  \eqref{eq:z-update} and  \eqref{eq:y-update}. \\
		\lnl{dn-7} Distribute $\zz^{k+1}$ and $\yy_i^{k+1}$ to all nodes. \\
		 Locally, on each node, compute spectral step sizes and penalty parameters as in \cite{xu2017adaptiveb}
	}
\end{algorithm}

Steps  \ref{dn:line:1}-\ref{dn:line:2} initialize the multipliers, $\yy$, and consensus vectors, $\zz$, to  zeros. 
In each iteration, Single Node Newton method, Algorithm \ref{alg:standalone-Newton}, is run with 
local $\xx_{i}$, $\yy_{i}$, and global $\zz$ vectors. Upon termination of 
Algorithm \ref{alg:standalone-Newton} at all nodes, resulting local Newton directions, $\xx_{i}^{k}$, 
are gathered at the master node, which generates the next iterates for vectors $\yy$ and $\zz$ using spectral
step sizes described in \cite{xu2017adaptiveb}. 
These steps are repeated until convergence. 

\begin{remark}
Note that in each ADMM iteration only \emph{one} round of 
communication is required (a ``gather'' and a ``scatter'' operation), 
which can be executed in $\mathcal{O}(\log (\mathcal{N}))$ time. Further, the application of the GPU-accelerated inexact Newton-CG Algorithm \ref{alg:standalone-Newton} at each node significantly speeds-up the local computation per epoch. The combined effect of these algorithmic choices contribute to the high overall efficiency of the proposed Newton-ADMM Algorithm \ref{alg:dist-Newton}, when applied to large datasets.
\end{remark}

\subsubsection{ADMM Residuals and Stopping Criteria }
The consensus problem \eqref{eq:dist-obj} can be solved by combining ADMM subproblems \eqref{eq:x-update}, \eqref{eq:y-update}, and \eqref{eq:z-update}. To monitor the convergence of ADMM, we can check the norm of primal and dual residuals, $\rr^k$ and $\dd^k$, which are defined as follows:
\begin{align}
\rr^{k} = \begin{bmatrix}
           \rr_{1}^k \\
           \vdots \\
          \rr_{\mathcal{N}}^k
         \end{bmatrix}, \dd^{k} = \begin{bmatrix}
           \dd_{1}^k \\
           \vdots \\
          \dd_{\mathcal{N}}^k
         \end{bmatrix} 
\end{align}
where $\forall i \in \{1,2,\dots,\mathcal{N}\}$,
\begin{align}
\rr_{i}^k = \zz^k - \xx_{i}^k, \dd_{i}^k =- \rho_i^{k}(\zz^{k}-\zz^{k-1})
\end{align}
 As $k\rightarrow\infty$,  $\zz^k \rightarrow\zz^*$ and  $\forall i, \xx_{i}^{k}\rightarrow\zz^*$. Therefore, the norm of primal and dual residuals, $|| \rr^k ||$ and $|| \dd^k ||$, converge to zero. In practice, we do not need the solution to high precision, thus ADMM can be terminated as $|| \rr_{i}^k || \leq \epsilon^{pri}$ and $|| \dd_{i}^k || \leq \epsilon^{dual}$ . Here, $\epsilon^{pri}$ and $\epsilon^{dual}$ can be chosen as:
\begin{flalign}
&\epsilon^{pri}=\sqrt{\mathcal{N}}\epsilon^{abs}+\epsilon^{rel}\max\{\sum_{i=1}^{\mathcal{N}}||\xx_{i}^k||^2,\mathcal{N}||\zz^{k}||^2\}\\
&\epsilon^{dual}=\sqrt{d}\epsilon^{abs}+\epsilon^{rel}\max\{\sum_{i=1}^{\mathcal{N}}||\yy_{i}^k||^2\}
\end{flalign}
The choice of absolute tolerance $\epsilon^{abs}$ depends on the chosen problem and the choice of relative tolerance $\epsilon^{rel}$ for the stopping criteria is, in practice, set to $10^{-3}$ or $10^{-4}$.

\subsection{Inexact Newton-CG Solver}
The optimization problem \eqref{eq:obj} is decomposed by ADMM framework into sub-problems \ref{eq:x-update}, \ref{eq:y-update}, and \ref{eq:z-update}. Among these sub-problems, only \ref{eq:x-update} does not have closed form solution. Thus, it is critical to find an iterative method that can produce high quality solutions with low computation cost. To this end, we develop an \textit{inexact Newton-CG solver} for  solving sub-problem \ref{eq:x-update}. Let the objective in Equation \ref{eq:x-update} be $\hat{f}(\xx)$, in each iteration; the gradient and Hessian are given by:\\
\begin{subequations}\label{eq:Newton_G_H}
\begin{equation}
\bgg(\xx) \triangleq  \sum_{j \in \mathcal{D}} \nabla \hat{f}_{j}(\xx) ,\label{Newton_G}
\end{equation}
    
\begin{equation}
\HH(\xx) \triangleq  \sum_{j \in \mathcal{D}} \nabla^{2} \hat{f}_{j}(\xx) . \label{Newton_H}
\end{equation}
\end{subequations}
At each iterate $ \xx^{(k)} $, using the corresponding Hessian, $\HH(\xx^{(k)})$,
and the gradient, $\bgg(\xx^{(k)})$, we consider \emph{inexact} Newton-type iterations of the form: 
\begin{subequations}
\begin{align}
\label{eq:update}
\xx^{(k+1)} = \xx^{(k)} + \alpha_{k} \pp_{k},
\end{align}
where $ \pp_{k} $ is a search direction satisfying:
\begin{align}
\label{eq:inexact}
\| \HH(\xx^{(k)})\pp_{k} + \bgg(\xx^{(k)})\| \leq \theta \|\bgg(\xx^{(k)})\|,
\end{align}
for some inexactness tolerance $ 0 < \theta < 1 $ and $\alpha_{k}$ is the largest $\alpha \leq 1$ such that
\begin{align}
F(\xx^{(k)} + \alpha \pp_{k}) \leq F(\xx^{(k)}) + \alpha \beta \pp_{k}^{T} \bgg(\xx^{(k)}),
\label{eq:armijo}
\end{align}
for some $\beta \in (0,1)$. 
\label{eq:inexact_Newton_itrs}	
\end{subequations}

Requirement \eqref{eq:armijo} is often referred to as Armijo-type
line-search \cite{nocedal2006numerical}. To compute the step-size, $\alpha$ in eq.  \eqref{eq:armijo}, 
~we use a \textit{backtracking} line search, as shown in algorithm \ref{alg:linesearch}.
This function takes parameters, $\alpha $ as initial step-size, which in our case is always set to 1, $ \beta < 1 $, $\pp$ is the Newton-direction, 
and the gradient vector is $\bgg$. The loop at line \ref{ls:line:3} is repeated until desired reduction
is achieved along the Newton-direction, $\pp$, by successively decreasing the step-size by a factor $ \gamma < 1$.

\begin{algorithm} [!htbp]
\caption{Line Search}
\label{alg:linesearch}
	\SetAlgoLined
	\SetKwInOut{Input}{Input}
	\SetKwInOut{Parameter}{Parameters}
	
	\Input{  \\
		$ \xx $ - Current point \\
	         $ \pp $ - Newton's direction \\
			 $F(.) $ - Function pointer\\
			 $\bgg(\xx)$ - Gradient vector
			 }
	\Parameter{\\
			$\alpha$ - Initial step size \\
			$0< \beta < 1$ - Sufficient descent constant\\
			$0< \gamma < 1$ - Back-tracking parameter \\
			$ i_{\max} $ - Maximum line search iterations 
			}
	$\alpha = 1$ \\
    $ i = 0 $ \\
	\While{ $ F ( \xx + \alpha  \pp) > F ( \xx ) + \alpha \beta  \pp^{T} \bgg(\xx) $ }{ \label{ls:line:3}
		\If{ $ i > i_{\max} $}{
			\lnl{ls:5} break \\
		}
		$ i = i + 1 $\\
		$ \alpha \leftarrow \gamma \alpha $	}
\label{alg-ls}
\end{algorithm}


Condition \eqref{eq:inexact} is the
$\theta$-relative error approximation of the exact solution to the linear system:
\begin{align}
\label{eq:exact}
\HH(\xx^{(k)}) \pp_{k} &= -\bgg(\xx^{(k)}),
\end{align} 
Note that in (strictly)
convex settings, where the Hessian matrix is symmetric positive definite (SPD),
conjugate gradient (CG) with early stopping can be used to obtain an approximate solution
to \eqref{eq:exact} satisfying \eqref{eq:inexact}. In \cite{roosta2019sub}, it has been shown that a mild value for $ \theta $, in the order of inverse of \emph{square-root of the condition number}, is sufficient to ensure that the convergence properties of the exact Newton's method are preserved. As a result, for ill-conditioned problems, an approximate solution to \eqref{eq:exact} using CG yields good performance, comparable to an exact update (see examples in Section \ref{sec:expt}). Putting all of these together, we obtain Algorithm \ref{alg:standalone-Newton}, which
is known to be globally linearly convergent, with problem-independent local convergence rate \cite{roosta2019sub}.

\begin{algorithm} [!htbp]
	\caption{Inexact Newton-type Method}
	\label{alg:standalone-Newton}
	\SetAlgoLined
	\SetKwInOut{Input}{Input}
	\SetKwInOut{Parameter}{Parameters}
	\Input{$\xx^{(0)}$}
	\Parameter{$0 < \beta, \theta < 1$}
	\ForEach{$k = 0,1,2,\ldots$}{
		Form $\bgg(\xx^{(k)})$ and $\HH(\xx^{(k)})$ as in \eqref{eq:Newton_G_H} \\
		\If{$\|\bgg(\xx^{(k)})\| < \epsilon$}{
			STOP
		}
		Update $\xx^{(k+1)}$ as in \eqref{eq:inexact_Newton_itrs}\\
	}
\end{algorithm}

\subsection{GPU Utilization and Communication Overhead}
\label{GPU-utils}
 Our proposed methods use Inexact Newton-type iterates, with a linear-quadratic convergence rate for strongly convex sub-problems \eqref{eq:x-update}. Besides theoretical performance guarantees, our Newton-type method has practical advantages over first order methods. In practice, mini-batch stochastic gradient descent is widely used over full-batch gradient descent and other methods. However, this method often requires a large number of epochs to achieve good generalization errors. Furthermore, the mini-batch update scheme results in significantly lower GPU occupancy (idle GPU cores because of smaller batch sizes). The number of CPU-GPU memory transfers per epoch for mini-batch SGD is $\frac{n}{m}$, where $n$ is the size of dataset, and $m$ is the size of mini-batch. Usually, $n >> 1$  and $m$ is typically between a hundred and a thousand. In contrast, Newton's method utilizes the complete dataset for computing Newton direction. Therefore, there is only one CPU-GPU memory transfer for computing Newton direction, which greatly increases utilization of the GPU for reasonably sized datasets. With a judicious mix of statistical methods and carefully formulated Hessian-vector operations, as discussed in Section \ref{sec:MVP}, we are able to transform this computation-heavy operation into an efficient and highly scalable GPU-accelerated operation with low memory overhead. For this reason, Newton's method, in general, is more amenable to high GPU utilization, due to higher concurrency and lower CPU-GPU data transfer cost, compared to SGD. Furthermore, in distributed implementations, Synchronous SGD induces a communication overhead of $\frac{nd\log(\mathcal{N})}{m\mathcal{N}}$, while ADMM only requires one round of communication. In solving high dimensional problems, where $d >> 1$, this overhead is significant when network bandwidth is limited. The gains from using Newton method and ADMM, along with low communication and high GPU utilization, as compared to first order methods, are demonstrated in our empirical results.

\subsection{Hessian Vector Product}
\label{sec:MVP}

Given a vector $\vv \in \mathbb{R}^{d}$, we
can compute the Hessian-vector product on the GPU without explicitly forming the Hessian.
For notational simplicity, define 
\begin{align*}
h(\aa,\xx) \defeq \frac{e^{\lin{\aa, \xx}-M(\xx)}}{\alpha(\aa)}, 
\end{align*}
where $M(\xx)$ and $ \alpha(\xx) $ were defined in eqs.~\eqref{eq:max_x} and~\eqref{eq:log_sum_trick}, respectively.
Now using matrices 
\begin{align*}
\stepcounter{equation}\tag{\theequation}\label{eq:matrix-a}
\V = \begin{bmatrix}
\lin{\aa_{1},\vv_{1}} & \lin{\aa_{1},\vv_{2}} & \dots  & \lin{\aa_{1},\vv_{C-1}} \\
\lin{\aa_{2},\vv_{1}} & \lin{\aa_{2},\vv_{2}} & \dots  & \lin{\aa_{2},\vv_{C-1}} \\
\vdots & \vdots & \ddots & \vdots \\
\lin{\aa_{n},\vv_{1}} & \lin{\aa_{n},\vv_{2}} & \dots  & \lin{\aa_{n},\vv_{(C-1)}}
\end{bmatrix}_{n \times (C-1)}, 
\end{align*}
and
\begin{align*}
\stepcounter{equation}\tag{\theequation}\label{eq:matrix-b}
\W = \begin{bmatrix}
h(\aa_{1},\xx_{1}) & h(\aa_{1},\xx_{2}) & \dots  & h(\aa_{1},\xx_{C-1}) \\
h(\aa_{2},\xx_{1}) & h(\aa_{2},\xx_{2}) & \dots  & h(\aa_{2},\xx_{C-1}) \\
\vdots & \vdots & \ddots & \vdots \\
h(\aa_{n},\xx_{1}) & h(\aa_{n},\xx_{2}) & \dots  & h(\aa_{n},\xx_{C-1}) \\
\end{bmatrix}_{n \times (C-1)},
\end{align*}
we compute
\begin{align*}
\stepcounter{equation}\tag{\theequation}\label{eq:matrix-c}
\vec{U} = \V \odot \W - \W \odot \Big( \big(\left(\V \odot \W\right) \ee \big) \ee^{T} \Big), 
\end{align*}
to get
\begin{align*}
\stepcounter{equation}\tag{\theequation}\label{eq:hessianvec}
\H\vv = \text{vec} \left( \A^{T} \vec{U} \right), 
\end{align*}
where $ \vv = [\vv_{1};\vv_{2};\ldots;\vv_{C-1}] \in \mathbb{R}^{d}$, $ \vv_{i} \in \mathbb{R}^{p}, i=1,2,\ldots,C-1 $, $\ee \in \mathbb{R}^{C-1}$ is a vector of all $1$'s, and each row of the matrix
$ \A \in \mathbb{R}^{n \times p}$ is a row vector corresponding to the $i^{th}$ data point,
i.e, $\A^{T} = \begin{bmatrix} \aa_{1}, \aa_{2}, \ldots, \aa_{n} \end{bmatrix}$.

Note that the memory overhead of our GPU-accelerated Newton-type method is determined by the dimensions of the matrices 
$ \vec{U} $, $ \V $ and $ \W $, which are determined by the local dataset size and number of classes in multi-class classification 
problem at hand. With reasonably sized GPU clusters this memory footprint can be easily managed for large datasets. 
This enables Newton-type method to scale to large problems inaccessible to traditional second-order methods. 

\begin{figure*}[ht]
\hspace*{-2.5cm}
    \begin{subfigure}[t]{0.5\textwidth}
        \includegraphics[width=180mm]{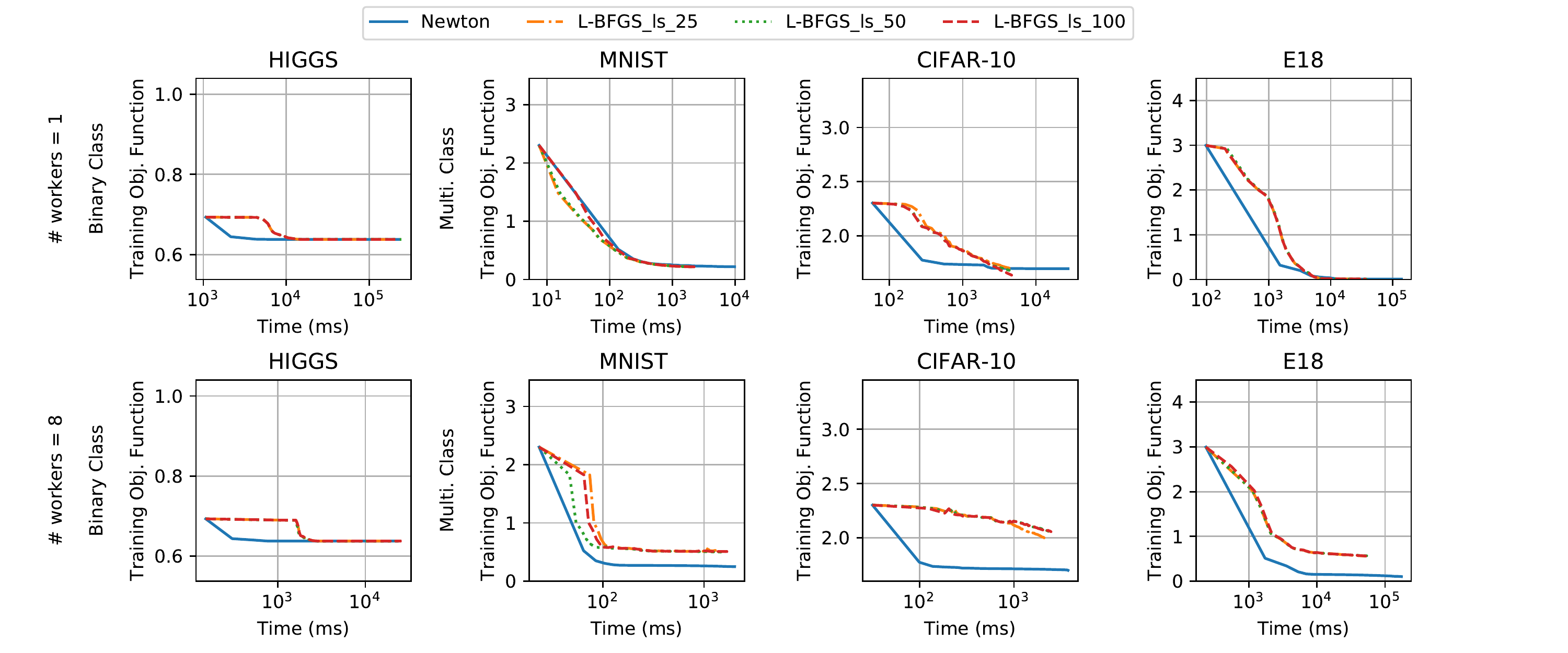}
    \end{subfigure}%
\caption{Training Objective function comparison over time for different choice of inner-solve for ADMM. For the inner solver, we compare the performance of Inexact Newton solver with L-BFGS (with history size 25, 50, 100). The step size of Inexact Newton method is chosen by linesearch following Armijo rule, whereas the step size of L-BFGS is chosen by linesearch satisfying Strong Wolfe condition. We can see that the per-iteration computation cost of L-BFGS is lower than Inexact Newton with the exception on HIGGS dataset. This is because L-BFGS is sensitive to the scale of step size so that more iterations of Strong Wolfe linesearch procedure are required to satisfy the curvature condition. In general, we observe that L-BFGS performs well on binary class problems, while the performance degrades on multiclass problems, when the number of compute nodes increases. }
\label{fig:ADMM_BFGS}
\end{figure*}

%% file: expt.tex
\section{Experimental Evaluation} 
\label{sec:expt}
In this section, we evaluate the performance of Newton-ADMM as compared with several state-of-the-art alternatives.
\begin{table}[h]
\small
\centering
\caption{Description of the datasets.}
\label{t1}
\begin{tabular}{|c|c|c|c|c|}
\hline
Classes&Dataset& Train Size & Test Size & Dims \\ \hline
2                 & HIGGS    & 10,000,000       & 1,000,000   & 28 \\ 
 10                & MNIST    & 60,000          & 10,000     & 784                \\
                10                & CIFAR-10 & 50,000          & 10,000     & 3,072               \\ 
                20                & E18      & 1,300,128         & 6,000      & 279,998             \\ \hline
\end{tabular}
\end{table}
\paragraph{\textbf{Experimental Setup and Data:}} All algorithms are implemented in PyTorch/0.3.0.post4 with Message Passing Interface (MPI) backend. We test performance of the methods on two hardware platforms. The first platform is a server with
 384 Intel Xeon Platinum 8168 processors and 8 Tesla P100 GPU cards. The second platform is a CentOS 7 cluster with 15 nodes with 100 Gbps Infiniband interconnect. Each node has 96GB RAM, two 12-Core Intel Xeon Gold processors, and 3 Tesla P100 GPU cards. We validate our proposed method using real-world datasets, described in Table~\ref{t1}, and compare with state-of-the-art first-order and second-order optimizers. These datasets are chosen to cover a wide range of problem characteristics (problem-conditioning, features, problem-size). MNIST is a widely used dataset for validation -- it is relatively well-conditioned. CIFAR-10 is 3.9x larger than MNIST and is relatively ill-conditioned. HIGGS\cite{baldi2014searching} is a low-dimensional dataset, however it is the largest (in terms of problem size). This dataset is easy to solve for our method, but is harder for first-order variants because of high communication overhead. The largest data set, E18  \footnote{    \href{https://support.10xgenomics.com/single-cell-gene-expression/datasets}{ E18 data set source.}     } , in terms of dimension and number of samples, is used to highlight  the scalability of our proposed method. 


\paragraph{\textbf{Newton-type method as a highly efficient subproblem solver for ADMM.}}
\label{LBFGS-ADMM}

We establish our (GPU-accelerated) Newton-type optimizer as a highly efficient inner solver for ADMM by comparing its performance against an ADMM-L-BFGS solver.  The per-iteration computation cost and memory footprint of L-BFGS is lower than our Newton-type method because of the rank-2 approximation of the Hessian matrix. This, however, comes at the cost of a worse convergence rate for L-BFGS. While Newton-type methods compute matrix vector products with the full Hessian, we use a Conjugate Gradient method with early stopping to solve the linear system, $ \HH \xx = -\bgg $. In our experiments we use no more than 10 CG iterations and a tolerance level of $ 10^{-3} $. The resulting \textit{Inexact} Newton-type method is GPU-accelerated, 
with an efficient implementation of Hessian-vector product. We show that, in practice, ADMM method suitably aided by efficient implementation of Newton-type subproblem solvers, yields significantly better results compared to the state-of-the-art. Furthermore, the use of true Hessian in our inexact solver, a second-order method, makes it resilient to problem ill-conditioning and immune to hyper-parameter tuning. These results are shown in Figure~\ref{fig:ADMM_BFGS}. We clearly notice that the performance gap between L-BFGS and Inexact Newton-type method becomes larger when number of compute nodes is increased. The only exception is on HIGGS dataset. This is because the dimension of the HIGGS datasets is only 28 and it is a binary classification problem. Consequently, the dimension of Hessian is significantly lower than among all the datasets. In this case, L-BFGS solver yields similar results compared to our Inexact Newton method. Most importantly, we note the following key results: (i) Inexact Newton yields performance improvements from 0 (MNIST) to 550\% (HIGGS and CIFAR-10) over L-BFGS on a single node; (ii) when using 8 compute nodes, the performance of L-BFGS-ADMM {\em never} catches up with that of Newton-ADMM (in terms of training objective function) in three of four benchmarks (MNIST, CIFAR-10, and E18), conclusively establishing the superiority of our proposed method over L-BFGS-ADMM.

\begin{figure*}[h]
\hspace*{-2.5cm}
    \begin{subfigure}[t]{0.5\textwidth}
        \includegraphics[width=170mm]{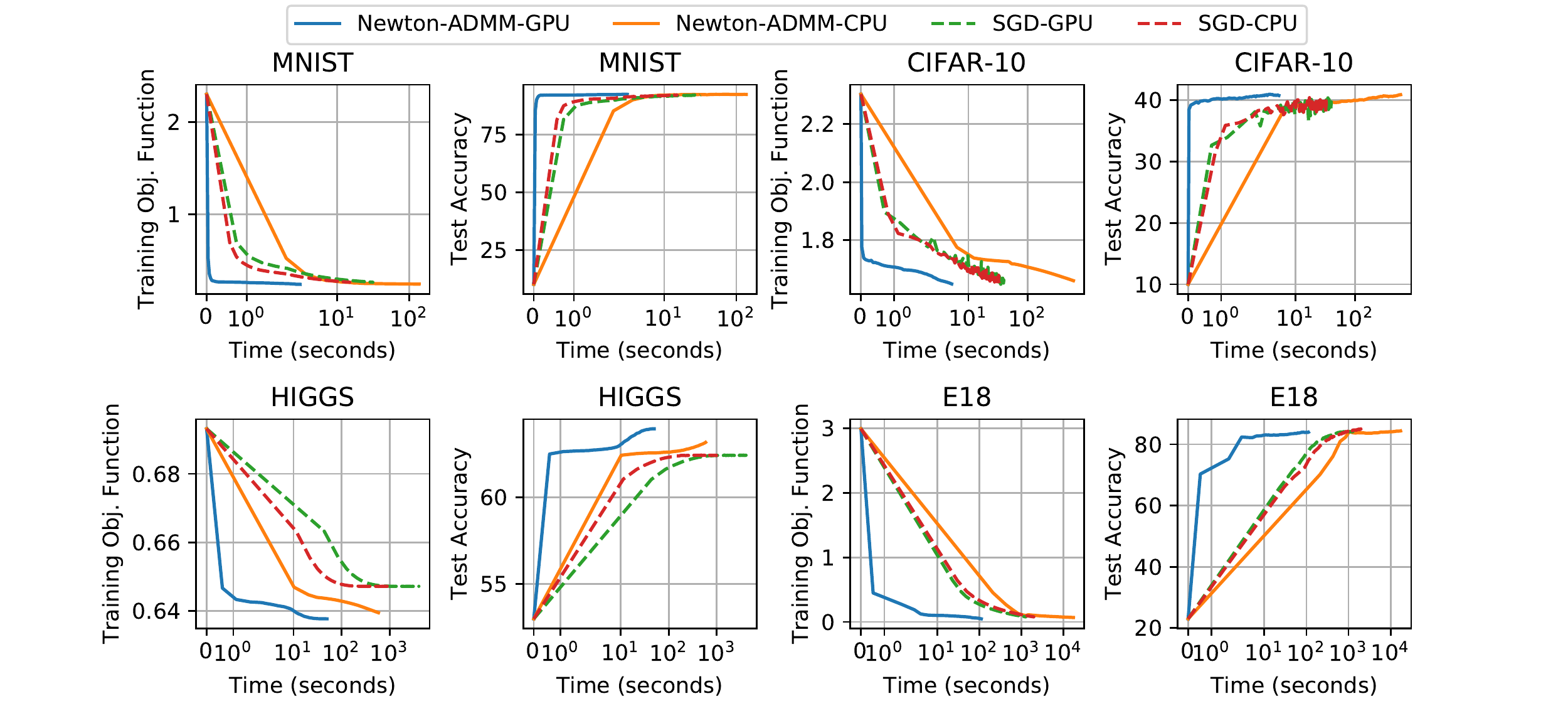}
    \end{subfigure}%
\caption{Training objective function and test accuracy as functions of time for Newton-ADMM and synchronous SGD, both with GPU enabled and GPU disabled, with 4 workers. Overall, Newton-ADMM favors GPUs, enjoys minimal communication overhead, and enjoys faster convergence compared to synchronous SGD. }
\label{fig:ADMM_SGD_Compare_1e-5}
\end{figure*}

\paragraph{\textbf{Comparison with Distributed First-order Methods.}} \label{sec:ADMM_vs_SGD}
 While the per-iteration cost of first-order methods is relatively low, they require larger numbers of iterations, increasing associated communication overhead, and CPU-GPU transactions, if GPUs are used (Please see detailed discussion in section \ref{GPU-utils}). In this experiment, we demonstrate that these drawbacks of first order methods are significant, both with the GPUs enabled and GPUs disabled. The results are shown in Figure \ref{fig:ADMM_SGD_Compare_1e-5}. Specifically, we note that GPU-accelerated Newton-ADMM method with minimal communication overhead yields significantly better results -- over an order of magnitude faster in most cases, when compared to synchronous SGD.


We present the ratio of CPU time to GPU time for Newton-ADMM and SGD in Table \ref{CPU-GPU-ratio}. We observe that for both Newton-ADMM and SGD, the CPU-GPU time ratio is proportional to the dimensions of datasets. For example, on the dataset with the lowest dimension (HIGGS), the CPU-GPU time ratio is the least for both  Newton-ADMM and SGD, whereas on the dataset with the highest dimension (E18), the CPU-GPU time ratios are the highest for both Newton-ADMM and SGD. In all cases, the use of GPUs results in highest speedup for Newton-ADMM. The gain in GPU utilization is compromised by large number of CPU-GPU memory transfers for SGD. As a result, SGD shows meaningful GPU acceleration only for the E18 dataset.

\begin{table}[h]
\caption{GPU Speedup for Newton-ADMM and SGD.}
\small
\label{CPU-GPU-ratio}
\centering
\begin{tabular}{|c|c|c|}
\hline
\textbf{\begin{tabular}[c]{@{}c@{}}CPU/GPU \\ Time Ratio\end{tabular}} & \textbf{\begin{tabular}[c]{@{}c@{}}Newton-ADMM\end{tabular}} & \textbf{SGD} \\ \hline
\textbf{MNIST}                                                         & 44.7345904                                                     & 0.47896507   \\ \hline
\textbf{CIFAR-10}                                                      & 112.670178                                                     & 0.8212862    \\ \hline
\textbf{HIGGS}                                                         & 11.842679                                                      & 0.26789652   \\ \hline
\textbf{E18}                                                           & 154.425688                                                     & 1.54673642   \\ \hline
\end{tabular}
\end{table}

Second, we observe that Newton-ADMM has much lower communication cost, compared to SGD. This can be observed from the Figure \ref{fig:ADMM_SGD_Compare_1e-5}. In all cases, SGD takes longer than Newton-ADMM with GPUs enabled. This is mainly because SGD requires a large number of gradient communications across nodes. As a result, we observe that Newton-ADMM is 4.9x, 6.3x, 22.6x, and 17.8x, times faster than SGD on MNIST, CIFAR-10, HIGGS, and E18 datasets, respectively.

Finally, we conclude that Newton-ADMM has superior convergence properties compared to SGD. This is demonstrated in Figure \ref{fig:ADMM_SGD_Compare_1e-5} for the HIGGS dataset. We observe that Newton-ADMM converges to low objective values in just few iterations. On the other hand, the objective value, even at 100-th epoch for SGD, is still higher than Newton-ADMM.

\paragraph{\textbf{Comparison with Distributed Second-order Methods.}}
We compare Newton-ADMM against DANE \cite{daneshmand2016dynaNewton}, AIDE \cite{reddi2016aide}, and GIANT \cite{wang2017giant}, which have been shown in recent results to perform well. 
In each iteration, DANE~\cite{daneshmand2016dynaNewton} requires an exact solution of its corresponding subproblem at each node. This constraint is relaxed in an inexact version of DANE, called InexactDANE \cite{reddi2016aide}, which uses SVRG \cite{johnson2013accelerating} to approximately solve the subproblems. Another version of DANE, called Accelerated Inexact
DanE (AIDE), proposes techniques for accelerating convergence, while still using InexactDANE to solve individual subproblems \cite{reddi2016aide}. However, using SVRG to solve subproblems is computationally inefficient due to its double loop formulation, with the outer loop requiring full gradient recalculation and several stochastic gradient calculations in inner loop.

\begin{figure}[h]
\centering
    \begin{subfigure}[t]{0.5\textwidth}
       \hspace*{-1.5cm} \includegraphics[width=85mm]{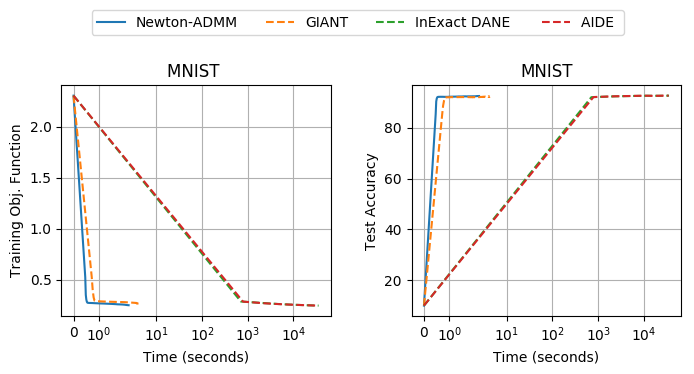}
    \end{subfigure}%
\caption{Training objective function and test accuracy comparison over time for Newton-ADMM, GIANT, InexactDANE, and AIDE on MNIST dataset with $\lambda=10^{-5}$. We run both Newton-ADMM and GIANT for 100 epochs. Since the computation times per epoch for InexactDANE and AIDE are high, we only run 10 epochs for these methods. We present details of hyperparameter settings in \ref{hyparameters}.}
\label{fig:DANE_weak_0_00001}
\end{figure}

Figure \ref{fig:DANE_weak_0_00001} shows the comparison between these methods on the MNIST dataset. Although InexactDANE and AIDE start at lower objective function values, the average epoch time compared to Newton-ADMM and GIANT is orders of magnitude higher (\textit{order of 1000x}). For instance, to reach an objective function value less than 0.25 on the MNIST dataset, Newton-ADMM takes only 2.4 seconds, whereas InexactDANE consumes \textit{an hour and a half}. Since InexactDANE and AIDE are significantly slower than Newton-ADMM and GIANT (on other datasets as well -- for which we do not show results here), we restrict our discussion of results on performance and scalability to Newton-ADMM and GIANT in the rest of this section.

\begin{figure*}[h]

    \begin{subfigure}[t]{0.5\textwidth}
    \centering
    \hspace*{-1.0cm}
    \hspace*{1.0cm}
        \includegraphics[width=120mm]{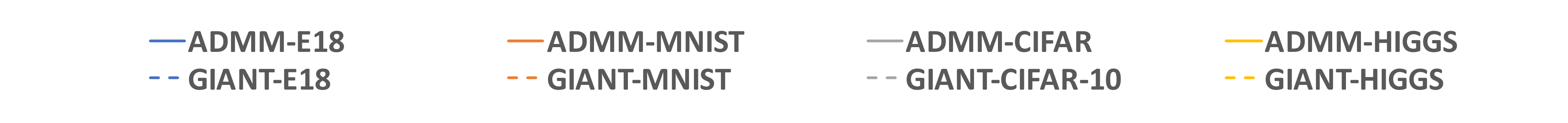}
    \end{subfigure}\\
    \begin{subfigure}[t]{0.5\textwidth}
    \hspace*{1.0cm}
    \includegraphics[width=100mm]{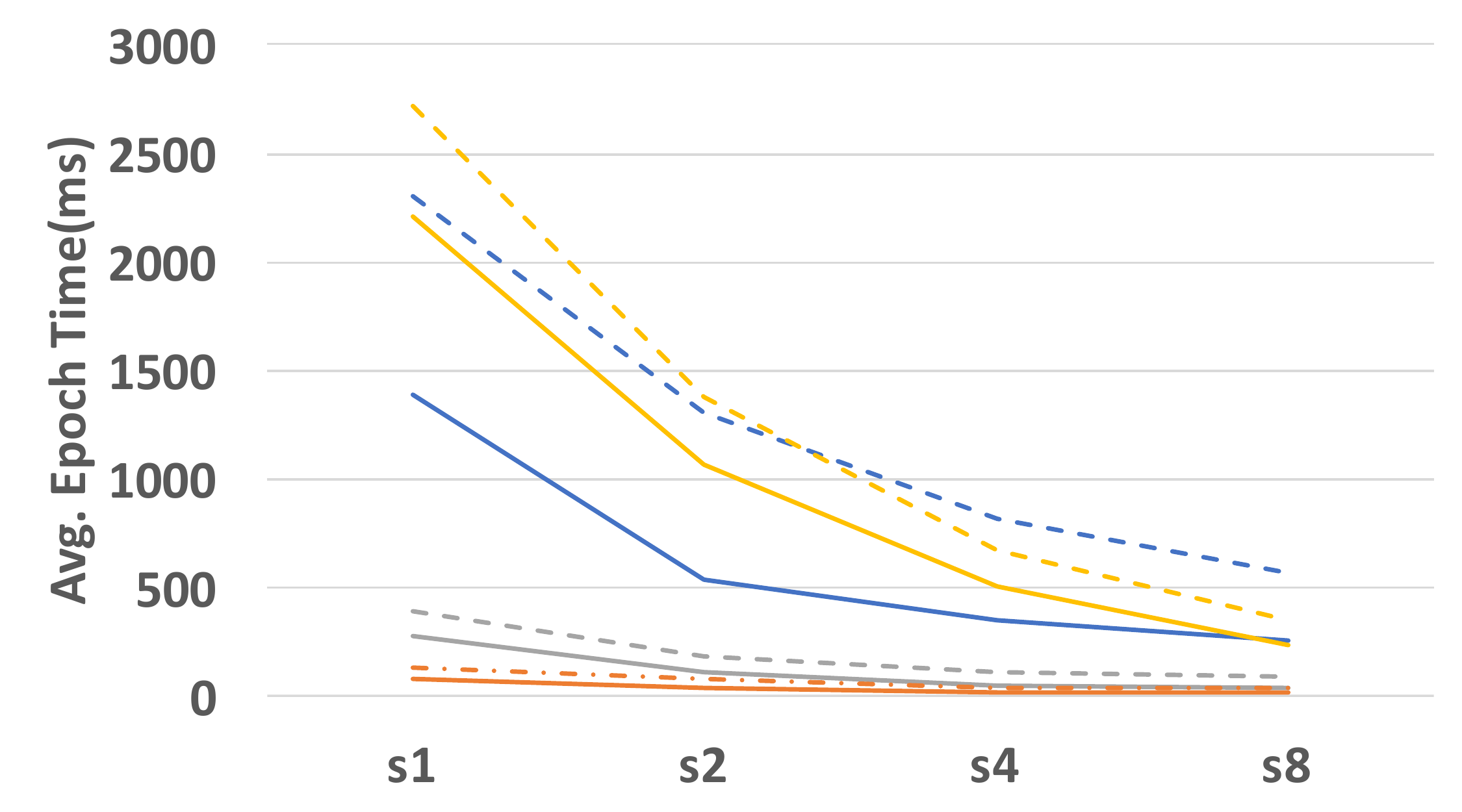}
    \end{subfigure}\\
    \begin{subfigure}[t]{0.5\textwidth}
    \hspace*{1.0cm}
    \includegraphics[width=100mm]{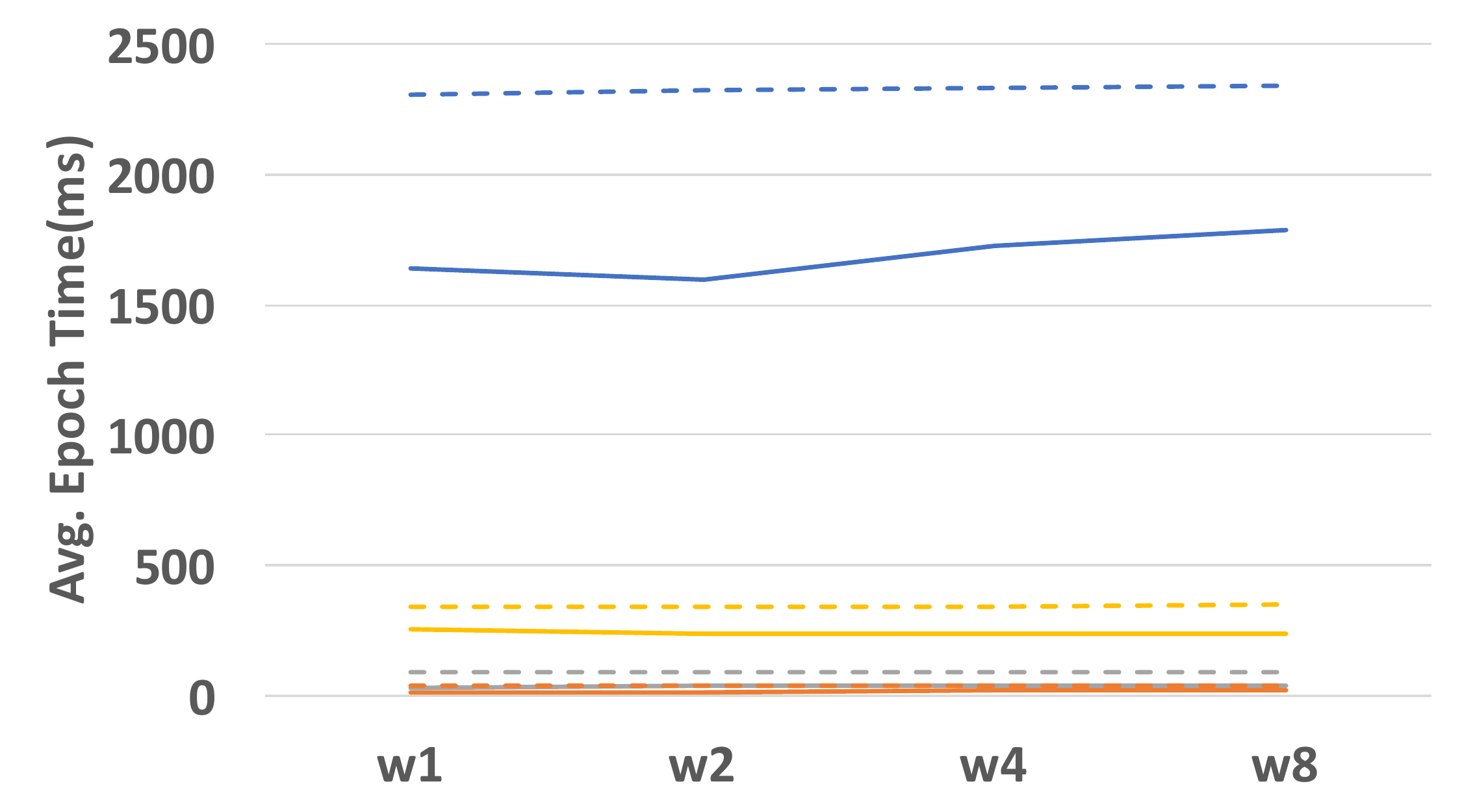}
    \end{subfigure}%
\caption{Avg. Epoch Time for Strong and Weak Scaling for Newton-ADMM and GIANT.}
\label{fig:avg_t}
\end{figure*}

\paragraph{\textbf{Scalability of Newton-ADMM.}} Figure \ref{fig:avg_t} presents strong- and weak-scaling results for Newton-ADMM and GIANT. In strong-scaling experiments, we keep the number of training samples constant, while increasing the number of workers, and for weak-scaling, the number of the training samples per node is kept constant. For strong-scaling, as number of workers increases, average epoch time for both Newton-ADMM and GIANT decreases. For all the datasets, as the number of workers is doubled, the average epoch time halved for both methods. For weak scaling, as the number of workers doubles, the average epoch time nearly remains constant for both methods. Both Newton-ADMM and GIANT use CG to compute Newton directions. However, compared to GIANT, Newton-ADMM has lower epoch times for the following reasons: first, to guarantee global convergence on non-quadratic problems, GIANT uses a globalization strategy based on line search. For this, the $i$-th worker computes the local objective function values $f_{\mathcal{D}_i}(\xx_{i}+\alpha\pp)$ for all $\alpha$'s in a pre-defined set of step-sizes $S=\{2^0,2^{-1},...,2^{-k}\}$, where $k$ is the maximum number of line search iterations. Thus, for each epoch, all workers need to compute a fixed number of objective function values. In contrast, Newton-ADMM performs line search only locally, allowing each worker to terminate line search before reaching the maximum number of line search iterations, and hence reducing the overhead of redundant computations. Second, Newton-ADMM only requires one round of messages per iteration, whereas GIANT needs three. Our experiments are performed on a Gigabit-interconnet cluster, where communication fabric is highly optimized. However, in environments with lower bandwidth and higher latency, we expect Newton-ADMM to perform significantly better compared to GIANT.

\begin{figure*}[t]
   \centering
    \begin{subfigure}[h]{0.5\textwidth}
        \hspace*{-3.5cm}
        \includegraphics[width=110mm]{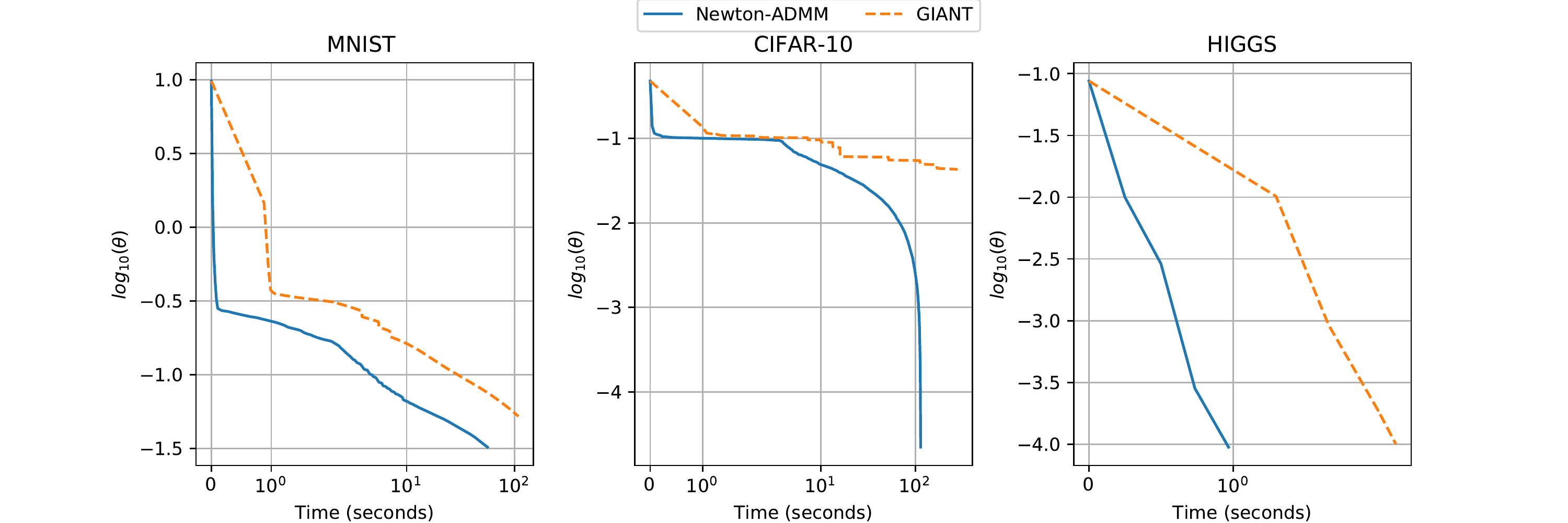}
        \caption{}
        \label{fig:ADMM_GIANT_t}
    \end{subfigure}%
    \hfill 
    \begin{subfigure}[h]{0.45\textwidth}
        \hspace*{-0.2cm}
        \includegraphics[width=90mm]{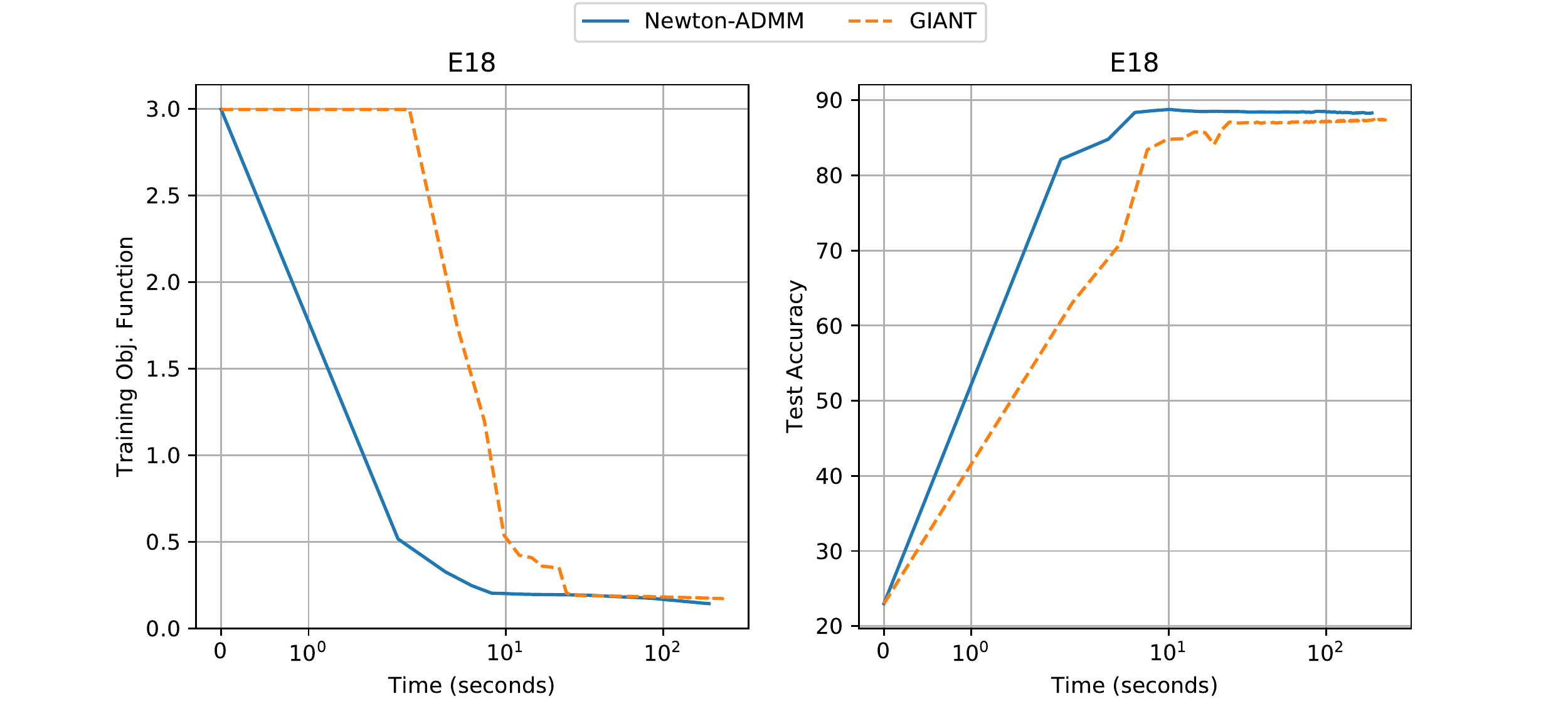}
        \caption{}
        \label{fig:ADMM_GIANT_E18}
    \end{subfigure}%
    \caption{Convergence and scalability performance of Newton-ADMM. Figure \ref{fig:ADMM_GIANT_t} shows $\log_{10}(\theta)$ as a function of time for Newton-ADMM and GIANT on MNIST, CIFAR-10, and HIGGS datasets. Newton-ADMM can reach lower $\theta$, given the same amount of time, compared to GIANT. Note that for the HIGGS dataset, both methods can reach low $\theta$ soon. Figure \ref{fig:ADMM_GIANT_E18} shows training objective function and test accuracy as function of time for Newton-ADMM and GIANT on E18 dataset using 32 nodes. We note that GIANT lingers at higher objective values in the initial iterations, while Newton-ADMM drops to lower objective values rapidly.}
    \label{fig:simulation}
\end{figure*}

We now compare the convergence of Newton-ADMM with GIANT in a distributed setting. Instead of comparing the test accuracy or objective value over time, we compare how close the objective value obtained from the solver is to the optimal objective value. Specifically, define $\theta = ({F(\xx^{k})-F(\xx^{*})})/{F(\xx^{*})}$, we measure $\theta$ as a function of time. (Here, $F(.)$ denotes the objective function, $\xx^{k}$ is the approximate solution obtained by the solver at the k-th iterate, and the ``optimal'' solution vector $\xx^{*}$ is obtained by running Newton's method on a single node to high precision).  Figure \ref{fig:ADMM_GIANT_t} shows $\theta$, in log scale, as a function of time for MNIST, CIFAR-10, and HIGGS using 8 compute nodes. From Figure \ref{fig:ADMM_GIANT_t}, we observe that, given the same amount of time, Newton-ADMM can reach lower $\theta$ in each case.
 We also measure the number of epochs taken by the solver to reach $\theta < 0.05$. Table \ref{t2} shows the number of epochs for Newton-ADMM and GIANT to reach $\theta < 0.05$ on 8 nodes. 
 \begin{table}[h]
\caption{Performance comparison of Newton-ADMM and GIANT -- we present the number of epochs for a solver to reach $\theta <0.05$. The speedup ratio is defined as the fraction of time taken by GIANT to achieve a specified value of $\theta$ to the corresponding time taken by Newton-ADMM on the same hardware platform.}
\centering
\small
\label{t2}
\begin{tabular}{|c|c|c|c|}
\hline
\textbf{}         & \textbf{\begin{tabular}[c]{@{}c@{}}NT-ADMM\\ Epochs\end{tabular}} & \textbf{\begin{tabular}[c]{@{}c@{}}GIANT\\ Epochs\end{tabular}} & \textbf{\begin{tabular}[c]{@{}c@{}}Speedup \\ \end{tabular}} \\ \hline
\textbf{MNIST}    & 252                                                                         & 1086                                                                   & 5.15                                                               \\ \hline
\textbf{CIFAR-10} & 1204                                                                         & 3215                                                                   & 11.14                                                              \\ \hline
\textbf{HIGGS}    & 1                                                                            & 1                                                                      & 1.35                                                               \\ \hline
\end{tabular}
\end{table} 

From Table \ref{t2}, we can see that Newton-ADMM converges to optimal solution significantly faster than GIANT.  Specifically, to reach $\theta \leq 0.05$, for the MNIST dataset, Newton-ADMM takes 252 epochs while GIANT takes 1086 epochs. For the CIFAR-10 dataset, Newton-ADMM takes 1204 epochs while GIANT takes 3215 epochs. The speed up ratio on MNIST and CIFAR-10 is 5.15 and 11.14, respectively. Both Newton-ADMM and GIANT behave well on HIGGS. It only takes 1 epoch for both solvers to reach $\theta \leq 0.05$. We note that the superior performance of these methods on HIGGS does not carry over to first-order methods.
 
 Finally, we stress that Newton-ADMM scales well on large datasets in large-scale distributed environments. From Figure \ref{fig:ADMM_GIANT_E18}, we note that Newton-ADMM takes significantly smaller amount of time to achieve lower objective values and higher test accuracy on E18 running on 32 compute nodes. 
 The large dimensionality of E18 (280K) highlights the memory- and compute-efficient formulation of our Hessian-vector products and subproblem solves on GPUs (please see section ~\ref{sec:MVP} for full GPU utilization characteristics of our solvers) -- the average epoch time for the E18 dataset on 32 nodes is only 1.98 seconds!

 \section{Conclusions and Future Directions}
\label{sec:conclusions}

We have developed a novel distributed Inexact Newton method based on a global consensus ADMM formulation. 
We compare our method with 
state-of-the-art optimization methods and show that our method has much lower distributed computing overhead, achieves
superior generalization errors, and has significantly lower epoch-times on standard benchmarks. 
We have also shown that our method can handle large datasets, while delivering sub-second epoch times -- establishing desirable scalability characteristics of our method.  Our results  establish Inexact Newton-ADMM as the new benchmark for 
performance of distributed optimization techniques.

%% file: conclusions.tex
 \section{Conclusions and Future Directions}
\label{sec:conclusions}

We have developed a novel distributed Inexact Newton method based on a global consensus ADMM formulation. 
We compare our method with 
state-of-the-art optimization methods and show that our method has much lower distributed overhead, achieves
superior generalization errors, and has significantly lower epoch-times on standard benchmarks. 
We have also shown that our method can handle large datasets, while delivering sub-second epoch times -- establishing desirable scalability characteristics of our method.  Our results  establish Inexact Newton-ADMM as the new benchmark for 
performance of distributed optimization techniques.

%% file: supplementary.tex
\section{Appendix}
In the appendix, we provide a detailed description of  parameter settings in all experiments.

\subsection{Algorithms Parameter Settings}
We generated all the experiment results using the following settings:
\label{hyparameters}
\begin{itemize}
	\item Synchronous SGD : we tune the step size from $10^{-4}$ to $10^4$ and select the best result to report.
	\item Newton-ADMM : We used 10 CG iterations along with $10^{-4}$ CG tolerance to compute Newton direction at each compute node. The step size was chosen by line search with 10 iterations.
	\item GIANT : The configurations for CG and linesearch are the same as the configurations use in Newton-ADMM. 
	\item Inexact DANE : we use learning rate $\eta=1.0$ and regularization term $\mu=0.0$ for solving subproblems as prescribed in \cite{daneshmand2016dynaNewton}. We set SVRG iterations to 100 and update frequency as $2n$, where $n$ is the number of local sample points. We run SVRG step size from the set $10^{-4}$ to $10^{4}$ in increments of 10 and select the best value to report. 
	\item AIDE : The configurations for SVRG is the same as the configurations used in Inexact DANE. As to the additional hyper-parameter introduced in AIDE, $\tau$, we also run $\tau$ from the set $10^{-4}$ to $10^{4}$ and select the best to report.
	\item Regularization parameter $\lambda$ : we used $\lambda = 10^{-5}$ for all the experiments.
\end{itemize}